\documentclass[a4paper,twoside]{article}

\usepackage{epsfig}
\usepackage{subcaption}
\usepackage{calc}
\usepackage{amssymb}
\usepackage{amstext}
\usepackage{amsmath}
\usepackage{amsthm}
\usepackage{multicol}
\usepackage{pslatex}
\usepackage{apalike}
\usepackage{hyperref}
\usepackage{SCITEPRESS}     

\begin{document}

\title{Point Cloud based Hierarchical Deep Odometry Estimation}

\author{\authorname{Farzan Erlik Nowruzi\sup{1},
		Dhanvin Kolhatkar\sup{1},
		Prince Kapoor\sup{2}
		and Robert Laganiere\sup{1,2}}
	\affiliation{\sup{1}School of Electrical Engineering and Computer Sciences, University of Ottawa, Canada}
	\affiliation{\sup{2}Sensorcortek Inc, Canada}
	\email{\{fnowr010, dkolh099\}@uottawa.ca, prince@sensorcortek.ai, laganier@eecs.uottawa.ca}
}

\keywords{Deep Learning, Lidar, Pointcloud, Odometry.}

\abstract{Processing point clouds using deep neural networks is still a challenging task. Most existing models focus on object detection and registration with deep neural networks using point clouds. In this paper, we propose a deep model that learns to estimate odometry in driving scenarios using point cloud data. The proposed model consumes raw point clouds in order to extract frame-to-frame odometry estimation through a hierarchical model architecture. Also, a local bundle adjustment variation of this model using LSTM layers is implemented. These two approaches are comprehensively evaluated and are compared against the state-of-the-art.}

\onecolumn \maketitle \normalsize \setcounter{footnote}{0} \vfill

\section{\uppercase{Introduction}}
\label{sec:introduction}
Autonomous vehicles should be able to operate in known or unknown environments. In order to navigate these environments, they have to be able to precisely localize themselves. A major issue in localization and mapping is caused by the tight coupling between these modules. We require highly precise maps for localization, while accurate localization is required to create precise maps. This inter-dependency has raised interest in methods that perform both tasks at the same time, termed Simultaneous Localization and Mapping (SLAM).

In an unknown environment, maps are not available for use as a priori of the environment model. A localization module is needed to infer the position on its own. One way to address this challenge is to estimate the amount of movement in between two individual observations and incrementally calculate the location of a sensor in the coordinate frame of the first observation. This is termed \textit{odometry}-based localization. We propose to build a novel method to tackle this problem by relying on the rich sensory data from lidar in an unknown environment.

To achieve this, we utilize the power of deep neural models to build the backbone of the odometry module. Deep models are not suited to perform tasks such as localization using point cloud data out of the box. In traditional methods, landmark extraction, data association, filtering and/or smoothing are used in a pipeline to get the final estimate. Our goal is to replace this pipeline with an end-to-end deep model.

Various traditional approaches~\cite{zhang2014loam}\cite{chen2020sloam} are proposed to perform scan-to-scan matching to extract odometry data. The majority of these models rely on using ICP and RANSAC to extract the registration. The majority of odometry estimation approaches utilize a temporal filtering stage that is classified as the Filtering or the Bundle Adjustment.
Filtering models summarize the observations in compact representations. This makes them lighter and faster than bundle adjustment methods that maintain a much larger set of observations and constantly refine their past and current predictions.

Deep neural models have revolutionized many aspects of the computer vision field \cite{VGG}\cite{inceptionv2}. Point-cloud processing is one of the challenging scenarios that deep models have a harder time expanding. This is due to the complexity in the scale and the unordered nature of the information representation in point-clouds. \cite{voxelnet}\cite{pointnet}\cite{pointnet++}\cite{flownet3d} have tackled this problem. Many of these approaches are designed to address the classification and segmentation tasks on point-clouds. A comprehensive review of these methods can be found in \cite{guo2019deep}. 
In this paper, we focus on the application of deep models to the odometry-based localization task. Using deep models for this purpose is a fairly new field and methods have not yet matured enough. Unlike the current models that extract image-like representation of the data, our model directly consumes point clouds.
We employ models proposed for segmentation and classification of point cloud data as our feature extraction backbone. 
More specifically, the input point-cloud data is processed using Siamese PointNet++ layers \cite{pointnet++}. It follows the same architecture as Flownet3D \cite{flownet3d} in order to extract the correlation between feature maps. The point-clouds used in Flownet3D are captured from a single object and consist of fewer points. The point-clouds used for odometry include a much larger number of points, shifts, moving objects and drastic changes in the environment. 
Flownet3D uses up-convolutions to extract the 3D flow between two point-clouds. Instead, we pass the features maps to fully connected layers to regress the rotation and translation parameters.

\section{\uppercase{Related Work}}
\label{litrev}

Traditional visual localization methods such as LSD-SLAM \cite{lsdslam} and ORB-SLAM \cite{orbslam} mainly rely on local features such as SIFT \cite{sift} and ORB \cite{orb} to detect the keypoints on camera images and track them through multiple frames. Their lidar-based counterparts, LOAM \cite{zhang2014loam} and SLOAM \cite{chen2020sloam}, utilize a similar processing framework with point-cloud data. The capabilities of these models are always limited by the repeatability of the hand-crafted features in consecutive frames. 

One major issue in the odometry challenge is to solve the \textit{data association} or \textit{registration} problem. RANSAC \cite{ransachg} like systems are commonly used to rule out the outliers. Once the data association is achieved, the transformation is estimated. 

\cite{zhang2014loam} introduces Lidar based Odometry and Mapping (LOAM), which is one of the most prominent works in this field. LOAM extracts key-points from lidar point-clouds and builds a voxel grid-based map. They dynamically switch between frame-to-model and frame-to-frame operation to simultaneously estimate odometry and build a map.
\cite{chen2020sloam} builds on the idea of LOAM by replacing the key-points with semantic objects.

The use of learned features has been shown to provide better results in many computer vision tasks in comparison to hand-crafted ones. Following the revolution of deep learning in image processing, \cite{chen2017deep} extracts deep learned features from images that are later used to perform the place recognition task.
One of the early deep networks for camera pose estimation is defined in \cite{posenet} and is termed PoseNet. It estimates the camera re-localization parameters in 6 Degree-of-Freedom (6-DoF) using a single image. In contrast with traditional methods that rely on Bag of Words (BoW) \cite{BoW}, this method only requires the network weights and is highly scalable for place recognition. However, as the network weights represent a map, each new location will require a new training.
\cite{brahmbhatt2018geometry} increases the performance of PoseNet by replacing the Euler angles with the log of unit-quaternions and incorporating odometry results from pre-existing methods. This parameterization of rotation only requires 3 values instead of 4 parameters of quaternions and avoids over parameterization. On the other hand, \cite{cho2019deeplo} shows that Euler angles are more stable than quaternion based loss in their study.
\cite{sattler2018benchmarking} provides a comparison of visual localization methods on multiple outdoor datasets with variable environmental conditions.

In recent years, there is an increasing interest in solving the odometry problem with deep learning models. One of the first works that directly tackles the visual odometry challenge through an end-to-end approach is proposed by \cite{deepvo}. Odometry is estimated by utilizing a \textit{9-layered} convolutional neural network with two LSTM layers at the end. 


Processing point cloud data is very different than processing images. Regularly, special tricks are applied on the point cloud to get an image-like representations and apply convolutional networks. 
\cite{li2019net} builds depth maps from lidar point-clouds and uses it to extract surface normals. Using multi-task learning, it tries to simultaneously estimate odometry and build attention masks for geometrically consistent locations through a Siamese model. Using their combined loss function, they achieve comparable results to traditional approaches.
\cite{cho2019deeplo} utilizes two Siamese networks; one with surface normals and the other with vertices. Features extracted from both Siamese branches are summed and passed to an odometry extraction network.

The deep neural models that take raw point-clouds as input are categorized in two groups: \textit{Voxel-based} and \textit{Point-based} methods. A voxel is a 3D partition in the 3D point-cloud space. 

Point-based approaches directly consume the points in the cloud. These methods rely on local region extraction techniques and symmetric functions to describe the selected points in each region.
\cite{pointnet} introduces the idea of approximating a symmetric function using a multi-layer perceptron in order to process unordered point cloud data. Convolutional filters are used on these features to perform 3D shape classification and object part segmentation. This work is expanded by PointNet++ \cite{pointnet++} to better extract features from local structures. This is achieved through the use of iterative farthest point sampling and grouping of the unordered points to hierarchically reduce their number and only maintain a more abstract representation of the original set.





Any deep learning approach that uses Siamese model for odometry relies on a matching layer that could be implemented in various forms. \cite{revaud2016deepmatching} introduces a new layer that hierarchically extracts the image features for dense matching that is used for flow estimation. Flownet \cite{ilg2017flownet} convolves features of one Siamese branch against the other and uses the results to explain the scene flow.

\section{\uppercase{Data Pre-processing}}
\label{dataproc}
We utilize KITTI an odometery dataset \cite{Geiger2012CVPR} in our experiments. In this section we describe the pre-processing steps to prepare the input data to our deep odometry model.

\subsection{Label Extraction}
The KITTI dataset employs global pose coordinates on the local frame. Pose information is provided from the view of the first frame as the center of the coordinate system. These, however, are not suitable labels for the frame-to-frame odometry estimation task. Frame-to-frame pose transformations are achieved through following formula:

\begin{equation}
\label{eq:gauss}
\begin{split}
X_{i+1} &= T_{i,i+1} \cdot X_{i}\\
T_{i,i+1} &= G_{0,i+1} \cdot G_{0,i}^{-1}
\end{split}
\end{equation}

$T_{i,i+1}$ is the local transformation between two coordinate centers $X_{i}$ and $X_{i+1}$. $G_{0,i}$ and $G_{0,i+1}$ are the global transformation from the first frame (center of the global coordinate frame) to the frames $i$ and $i+1$. 



\subsection{Point-Cloud Sampling}
The point-clouds in the KITTI dataset consist of $100k$ points per frame. This is a huge set of points. Given that lidar observations become less reliable at longer distances, we remove any point farther than $50m$ from the center in the $x$ and $z$ dimensions. There is still a large number of points remaining after this step. To reduce these points, we use the farthest point sampling strategy \cite{eldar1997farthest} to sample only $12k$ and $6k$ points that are later used in our experiments.

Point-clouds collected in driving environments usually contain a large amount of ground points, which are not useful to extract motion information and are usually discarded as a pre-processing step \cite{ushani2017learning}\cite{flownet3d}. However, ground as a flat surface can provide cues regarding the pitch and roll orientations of the vehicle, which is valuable for the estimation of 6-DoF odometry. We choose a $75\%-25\%$ sampling ratio for non-ground and ground points, in an attempt to balance the size of the data and the accuracy of the system.

\subsection{Dataset Augmentation}

Another major issue with using deep learning models for odometry results from the highly imbalanced nature of the labels: the majority of roads are designed as straight lines, with only a few turns made in long drives. 
Deep learning models are highly dependent on datasets being as complete as possible, which is not the case in these scenarios.

On the dataset side, we employ over-sampling of the minority labels to address this challenge. 
As a result, models can learn the dynamics of the environment while assigning the required attention to minority transformations.

In 6-DoF motion, there are three rotation and three translation parameters that need to be estimated. Rotation components include $\alpha$, $\beta$, and $\gamma$ as the pitch, yaw, and roll, and translation components consist of motion in $x$, $y$, and $z$ axes. There are various methods that perform over-sampling - e.g.  repetition, or Synthetic Minority Over-Sampling Technique \cite{SMOTE}. 

In order to achieve a more balanced dataset, we use the repetition method based on the amount of divergence from the average odometry value. To augment the dataset, we first calculate the average and standard deviation for the rotation and translation component norms individually. 
To decide on which samples to use, we follow 3 rules:

\begin{itemize}
	\item Translation: if $x_t < |\mu_t-a\sigma_t|$ then augment with $T$.
	\item Rotation: if $x_r < |\mu_r-a\sigma_r|$ then augment with $T$.
	\item If both translation and rotation rules are satisfied perform another augmentation run with $x$.
\end{itemize}

Subscripts $r$ and $t$ represent the rotation and translation components, respectively.
$x$ is the input norm, $\mu$ is the average value, and $\sigma$ is the standard deviation for the corresponding input $T$. $a$ is the ratio that is used for the range interval.
 
Once a point pair is chosen to be repeated, number of copies $N_x$ is calculated using the following formula.
\begin{equation}
\label{eq:repetition}
N_x = \lceil {2}^{\frac{x-\mu}{a\cdot\sigma}}\cdot\frac{1}{D} \rceil
\end{equation}
$D$ is a divisor value that explicitly controls the magnitude of the repetitions and is set to $4$.

In this way, we add approximately $10000$ samples for rotation, $6000$ samples for translation, and $2000$ samples for both.

In the majority of driving scenarios, the cars are in motion and are seldom stopped. To address this, we repeat the identity transformation with a random probability of $10\%$ on the dataset. This results in the addition of approximately $2000$ samples to the training set.


\begin{figure}
	\begin{center}
		\includegraphics[width=\linewidth]{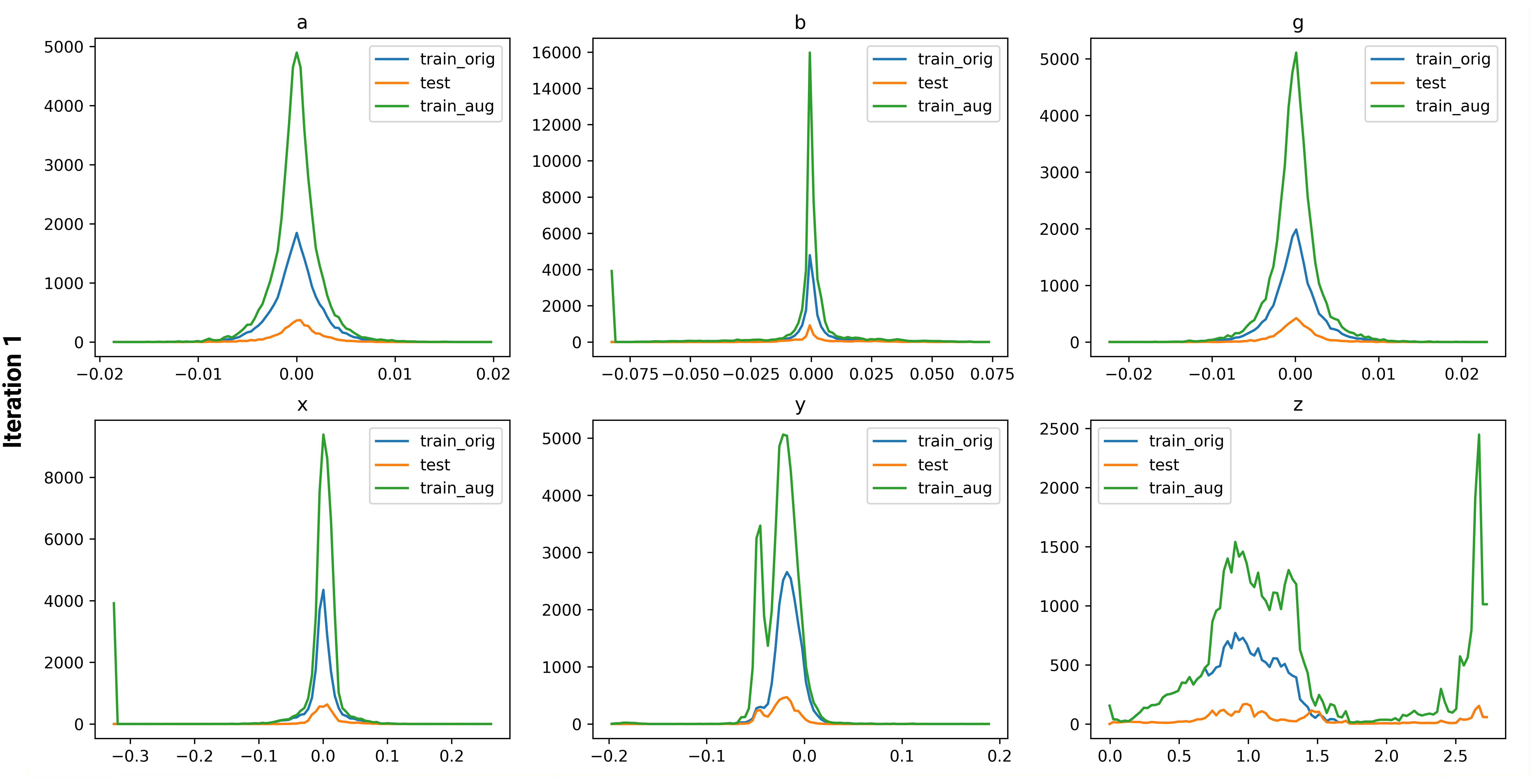}
	\end{center}
	\caption{Histograms of the consecutive transformation parameters before and after data augmentation.}
	\label{fig:data_distrib}
\end{figure}

Figure \ref{fig:data_distrib} compares the distribution of data points before and after augmentation. The imbalance in specific components is not completely removed, but is less than in the original dataset. This is especially the case for yaw $(b)$ and $z$ components that have a major effect on the accuracy of odometry. 
It is worth mentioning that, as all these components are correlated with each other, completely removing the imbalance only using this data is an impossibly challenging task.


\section{\uppercase{Model Architecture}}
\label{model}

\subsection{Proposed Core Model}
We rely on PointNet++ layers \cite{pointnet++} to build our model. Similar to their model, we propose a Siamese network that is able to regress the transformation between two point-clouds. Instead of passing the whole point-cloud to a PointNet feature extraction layer, we divide the inputs into two groups; one for ground points, and one for non-ground points. 
For ground points, we only use a single PointNet layer with a grouping distance threshold of $4$ that outputs $400$ points along with their descriptors. Descriptors are generated using 3 consecutive multi-layer perceptrons (MLP) of size $\left(64,96,128\right)$.
For non-ground points, there are two layers that subsequently use grouping distances of $0.5$ and $1$. $1500$ points are produced in the first layer and they are summarized to $800$ points in the second layer. Both of these layers use 3 MLPs where the first one consists of $(64,80,96)$ and the second one has layers of size $(112,128,128)$.
The distance metric used to group ground points is larger than the non-ground ones. This is due to the harsher sampling performed on the ground points that has resulted in larger distances between the points.

As the PointNet++ layers use farthest point sampling internally, we keep the ground and non-ground features separate for the feature extraction layer. The final outputs of ground and non-ground segments both have a dimensionality of $128$. Both feature maps are concatenated along the points dimension building a feature map of size $1200\times128$. At this stage, features from each frame are passed to the flow embedding layer of \cite{flownet3d}. We use the \textit{cosine} distance metric to correlate the features of each frame to each other. For further future extraction in this layer, we use the MLP with $(128,128,128)$ width. Through some experimentation, we found using nearest neighbor with $k=10$ gave the best results at this step. The final output shape of this layer is $1200\times128$.

Once the embedding between two frames is calculated, we run another feature extraction layer with radius of $1$ and MLPs of size $128,96,64$. This layer is also responsible for reducing the number of feature points that results in a feature map of shape $300\times64$. All of the layers up to this stage include batch normalization \cite{batchNorm} after each convolution. 

The resulting 2D feature map is flattened and a drop-out layer with keep probability of $0.6$ is applied on it. In order to extract the rotation and translation parameters, we use two independent fully connected layers of width $128$ and $3$.

\begin{figure*}
	\begin{center}
		\includegraphics[width=0.9\linewidth]{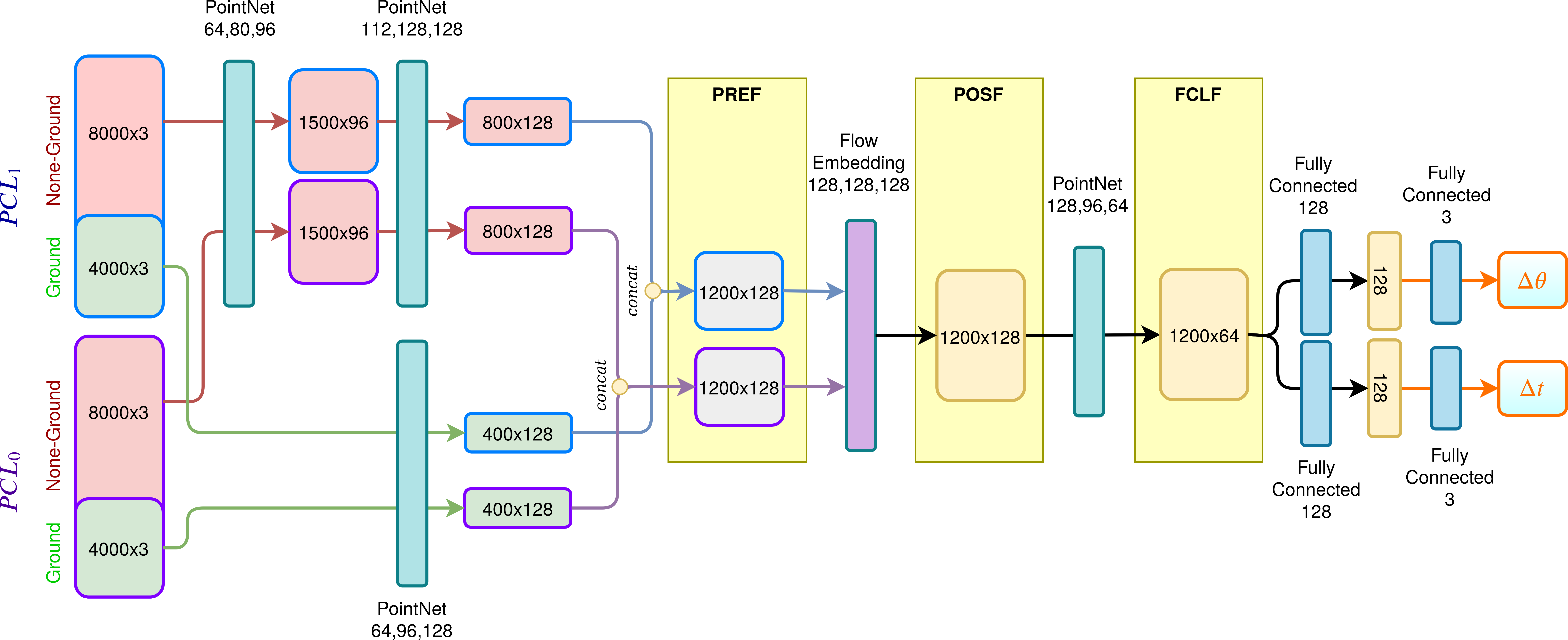}
	\end{center}
	\caption{Proposed core registration model architecture to process two consecutive frames and extract transformation parameters in between them.}
	\label{fig:NetCoreArch}
\end{figure*}

The final outputs of size $3$ are concatenated to build the 6-DoF Euler transformation parameters. Figure \ref{fig:NetCoreArch} shows the architecture of this network.

\subsection{Hierarchical Registration Model}

\begin{figure}
	\begin{center}
		\includegraphics[width=\linewidth]{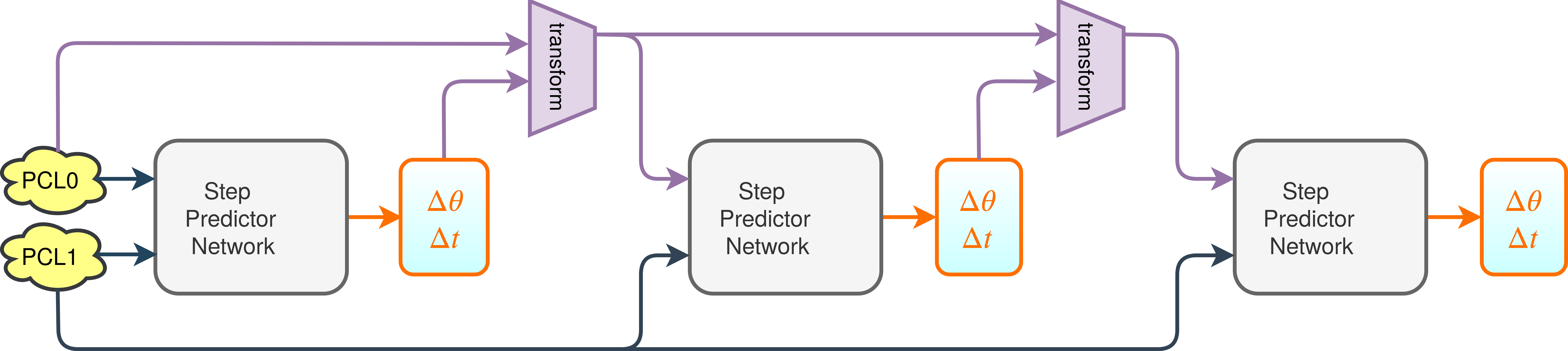}
	\end{center}
	\caption{Progressive Prediction. In the first iteration, $\Delta\theta$ and $\Delta t$ is estimated using $\mathrm{PCL}_{0}$ and $\mathrm{PCL}_{1}$. The source point cloud $\mathrm{PCL}_{0}$ is transformed to $\mathrm{PCL}_{0'}$. The residual transformation between $\mathrm{PCL}_{1}$ and $X_{0'}$ is calculated and used to estimate the next iteration of the model.}
	\label{fig:prog_odom_step}
\end{figure}

In traditional literature (optimization or RANSAC models) \cite{kitt2010visual}\cite{badino2013visual}, an odometry prediction is refined quickly through multiple iterations. We argue that the same could be applied in the case of deep learning-based odometry estimation models. Inspired by a hierarchical homography network \cite{now2017}, we use multiple layers of the same network to train on the residuals of previous predictions. The new ground truth is calculated by multiplying the ground truth from the previous iteration by the inverse of its prediction. In this way, each network reduces the dynamic range of error from the previous models and successively achieves a better result. Figure \ref{fig:prog_odom_step} shows this process.

The higher accuracy comes with the cost of increased train and test times. The key element in this approach is to keep the computational complexity as low as possible for each module in order to satisfy real-time processing requirements for the odometry task. 


\subsection{Temporal Filtering}
The goal of the odometry model is to smooth the effect of errors caused by the registration network. Our proposed model requires a large amount of memory due to its mid-level feature representation. Adding the memory requirements that the temporal model imposes, training quickly becomes a challenge for the system. Furthermore, training the core registration network is already a difficult task. Extra parametrization from the LSTM model makes an already difficult task an even more challenging one.

To alleviate these issues we propose a two stage training approach that breaks the initial feature extractor and the temporal filter into two disjoint models. Once the core registration network is trained, mid-level features are extracted and used as inputs to train the temporal model.

\begin{figure}
	\begin{center}
		\includegraphics[width=\linewidth]{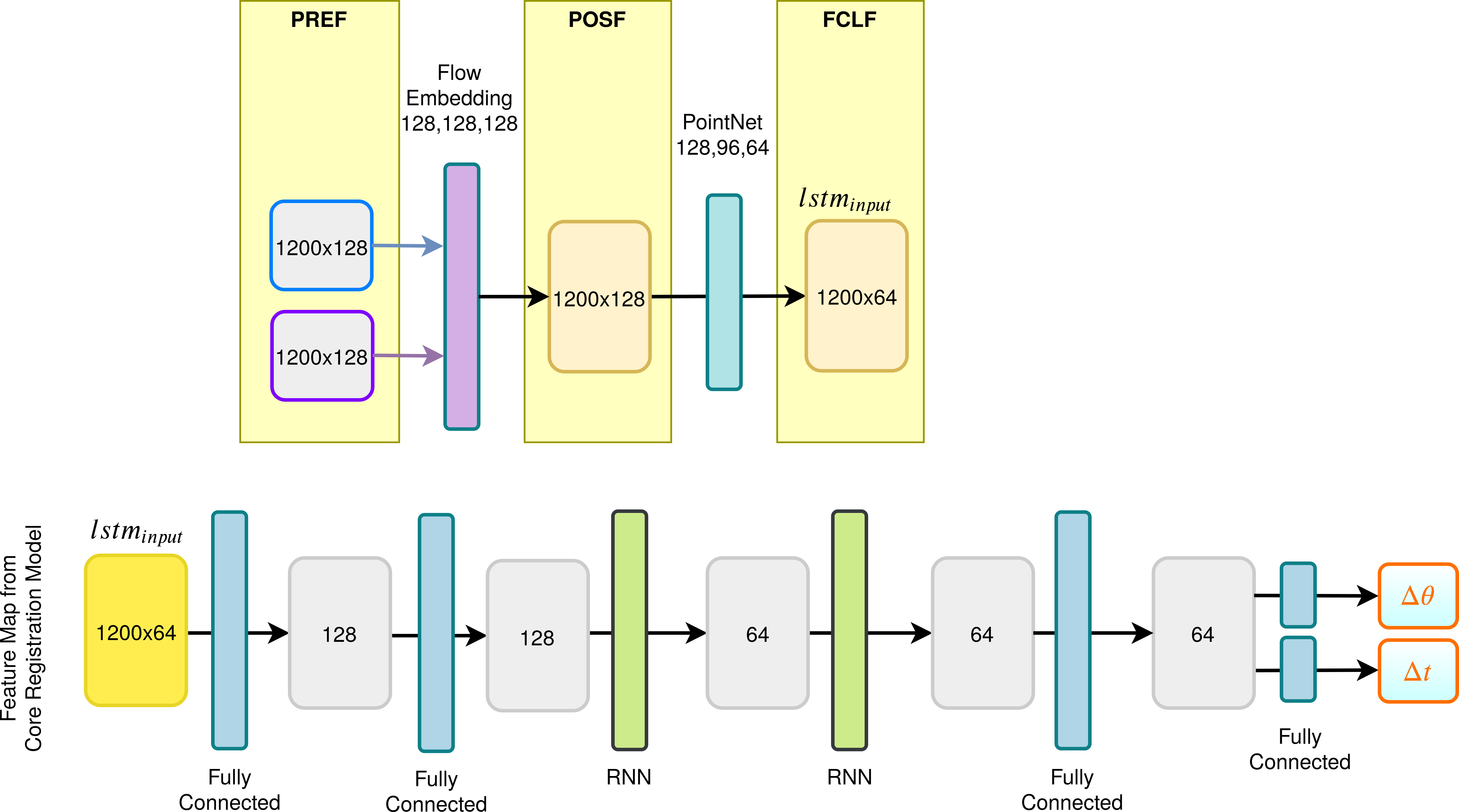}
	\end{center}
	\caption{Temporal model along with the three different input levels. FCLF is the natural input state of the temporal model. PREF and POSF require additional layers from the registration model to be included and re-learned inside the temporal model.}
	\label{fig:temporal}
\end{figure}

The base of our proposed temporal model consists of two fully connected layers of width $64$ and $128$. It is followed by two bidirectional LSTM with $64$ hidden layers with a \textit{soft\_sign} activation function. The output of the LSTM is fed to a fully connected layer with a size of 64. Finally, in the output layer, we have two fully connected networks with 3 nodes each that estimate the rotation and translation, individually. Figure \ref{fig:temporal} shows this model.

We employ drop-out and batch-normalization after each layer. For the LSTM model, a temporal window of $5$ frames is utilized.


\subsection{Loss Function}
As explained in Section \ref{dataproc}, the odometry labels suffer from data imbalance problem. The majority of the instances in the dataset follow a straight line, with only a few of them constituting the turns. To tackle these challenges, we incorporate measures in our loss function.

Using the naive \textit{L2-norm} diminishes the effect of instances with larger errors, as most of the batches result in smaller errors. It is required to make the model more sensitive towards larger errors. This is achieved through the use of online-hard-example-mining (OHEM) loss \cite{ohem}. In OHEM, only the top $k$ samples with highest loss values are used to calculate the final loss. However, once this value is set it does not change during the train. This could result in a training session that has high fluctuations in loss values. To address this, we implement an adaptive version of OHEM loss, that increases the number of top $k$ after certain epochs. The model initially focuses on the hardest examples, before its attention shifts towards all of the examples. This provides the hardest examples that are less frequent with a chance to drive the network towards the global minimum.

Another aspect of the loss function is to analyze the effects of errors in each component of the label. Rotation and translation components are separately extracted from the model. To enable the model's capability of adapting to each component, instead of using the naive approach, we employ a weighting mechanism.
To introduce uncertainty in our loss function, we consider the $log$ of normal distribution and phrase it as a minimization problem. In odometry, there are 6 parameters to be learnt that represent two components. We use the same weight for parameters related to each component.
In order to practically implement this loss function, we replace the standard deviation $\sigma$ of normal distribution with $\exp(w)$ that ensures the learned uncertainty is represented by a positive value. This results in the following final loss function.

\begin{equation}
\begin{split}
l_w = & \exp(-w_r)\|x_r-\hat{x_r}\|+w_r+\\
& \exp(-w_t)\|x_t-\hat{x_t}\|+w_t
\end{split}
\end{equation}

$x_t$ and $x_r$ are predicted translation and rotation variables. $\hat{x_t}$ and $\hat{x_r}$ define the ground-truth. $w_r$ and $w_t$ are the trainable weighting parameters, and $l_w$ defines the final loss. 





\section{\uppercase{Experiments}}
\label{experiments}
Estimating 6-DoF odometry using point-clouds is a challenging task. All the points are down sampled as described in section \ref{dataproc}. We first use the registration model to extract the transformation between two frames. This could be repeated for all the point-cloud pairs to generate the odometry trace. However, the additive nature of the noise quickly drifts the trace away from the ground truth. Due to this fact, there is no single metric that can explain all the various failures in the system.
We evaluate various aspects of the model using KITTI odometry metrics; Absolute Trajectory Error (ATE), Sequence Translation Drift Percentage, Sequence Rotation error per 100 meters, Relative Rotation Error (RRE) and Relative Translation Error (RTE)





RTE and RRE are the most important metrics as they evaluate the effectiveness of model for frame-to-frame estimation, independent of the trajectory. This is the main target that model is trained for.

\subsection{Hierarchical Model}

\begin{figure}
	\begin{center}
		\includegraphics[width=\linewidth]{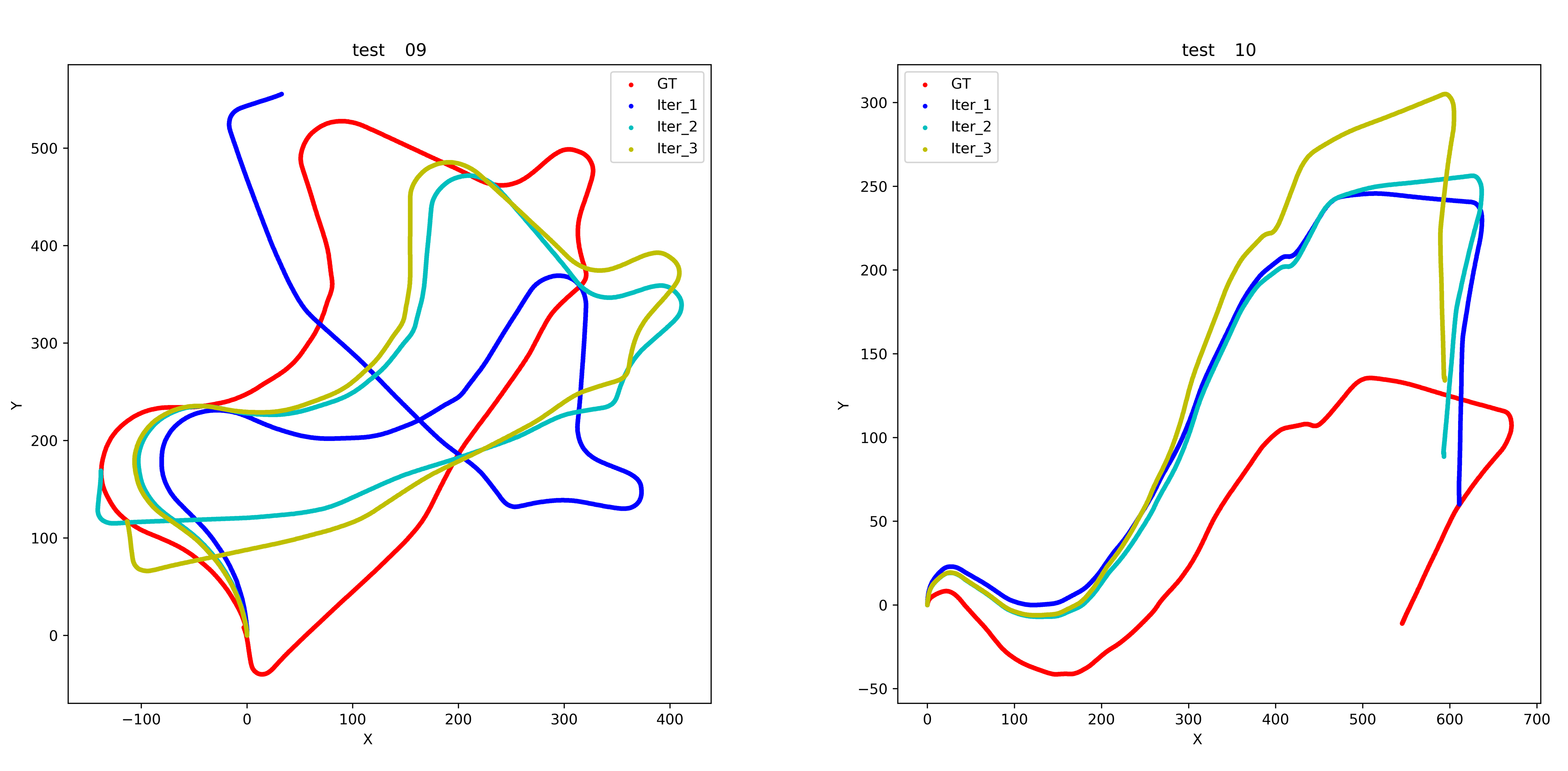}
	\end{center}
	\caption{\label{fig:trace}Estimated odometry trace for KITTI sequences using \textit{8k4k\_comp} model. Sequences 00-08 are used to train, and sequences 09-10 are used as test data for the model.}
\end{figure}

\begin{figure}
	\begin{center}
		\includegraphics[width=0.45\linewidth]{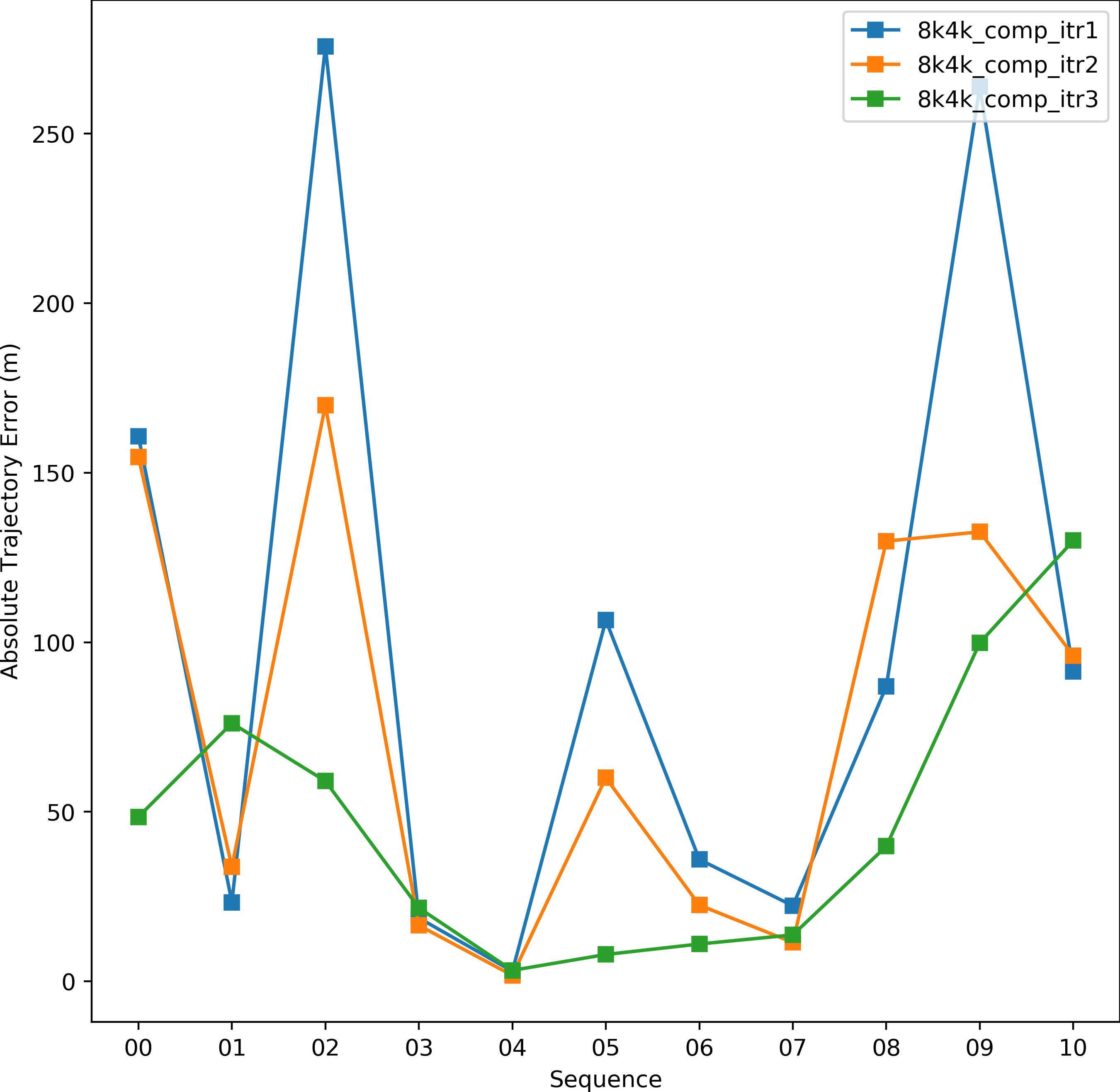}
		\\\vspace{0.2cm}
		\includegraphics[width=0.45\linewidth]{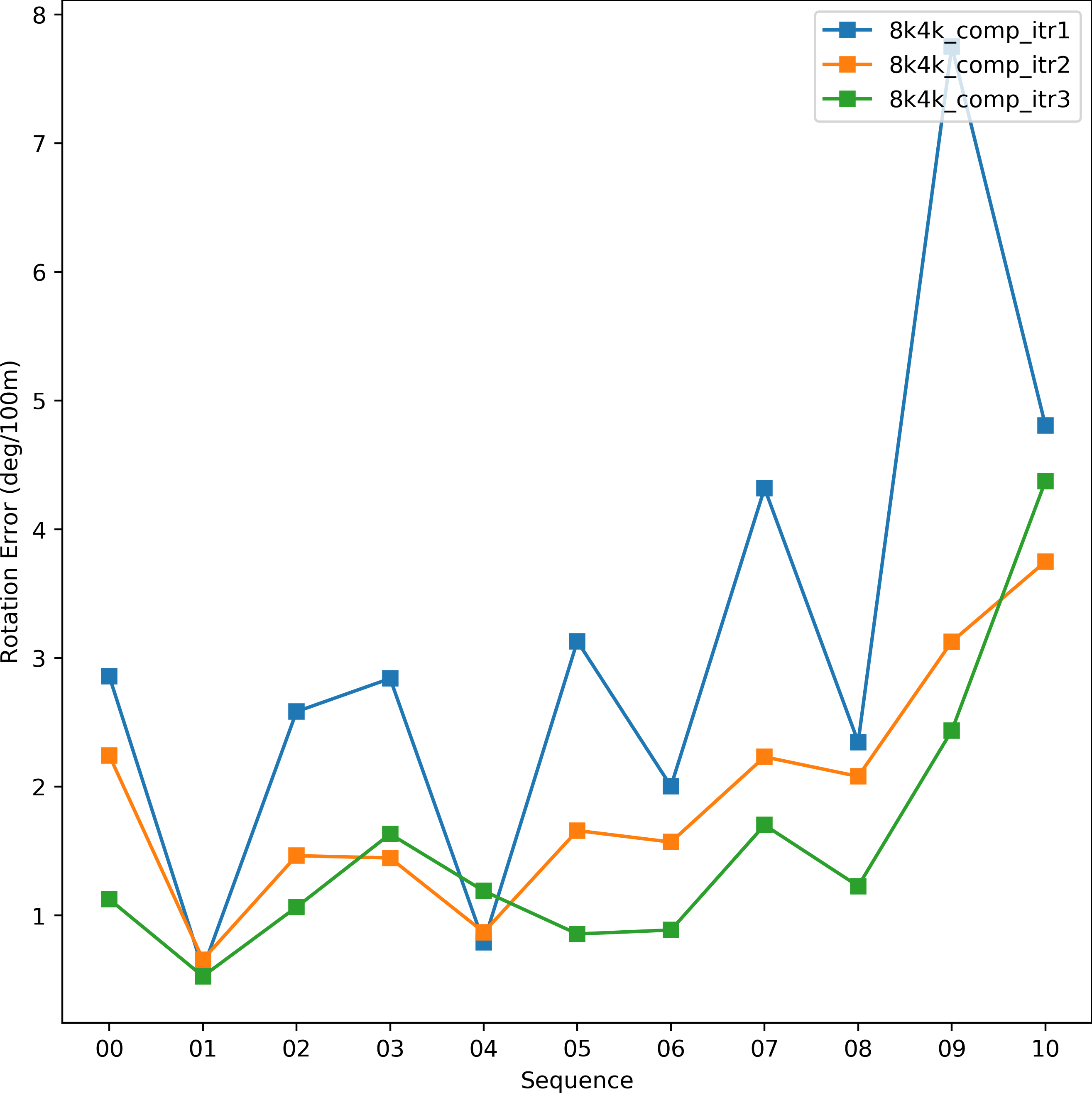}
		\includegraphics[width=0.45\linewidth]{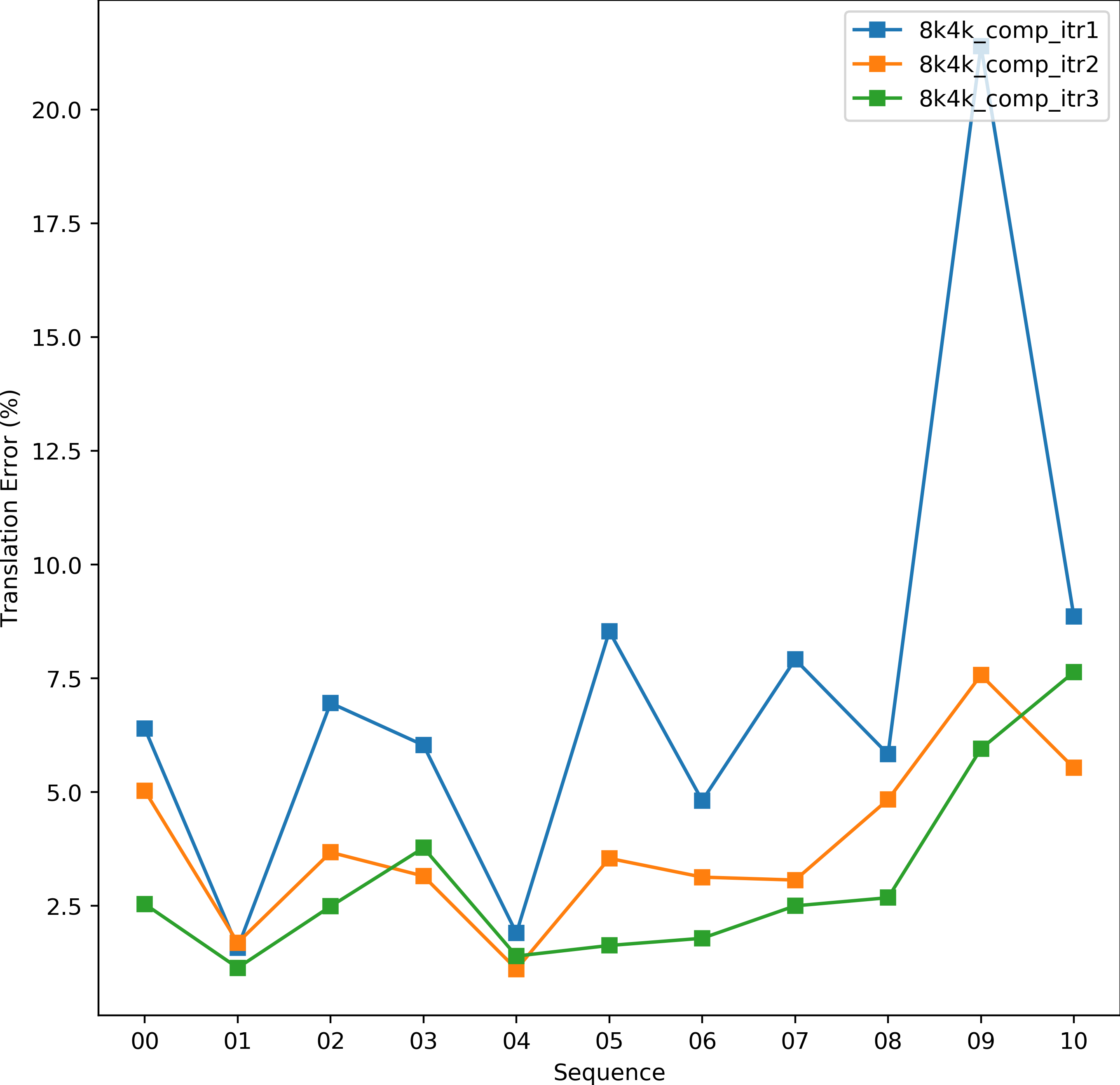}
		\\\vspace{0.2cm}
		\includegraphics[width=0.45\linewidth]{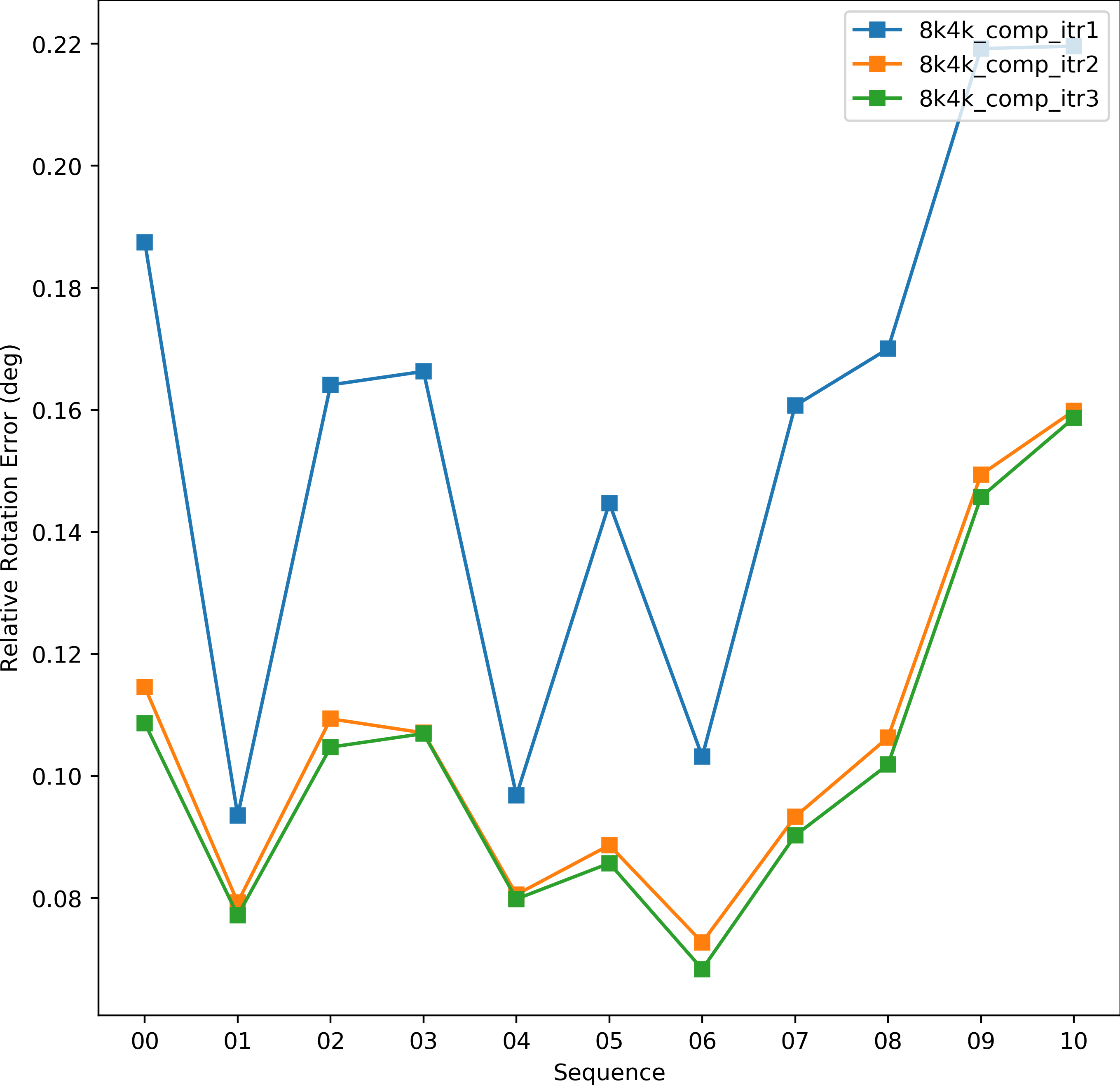}
		\includegraphics[width=0.45\linewidth]{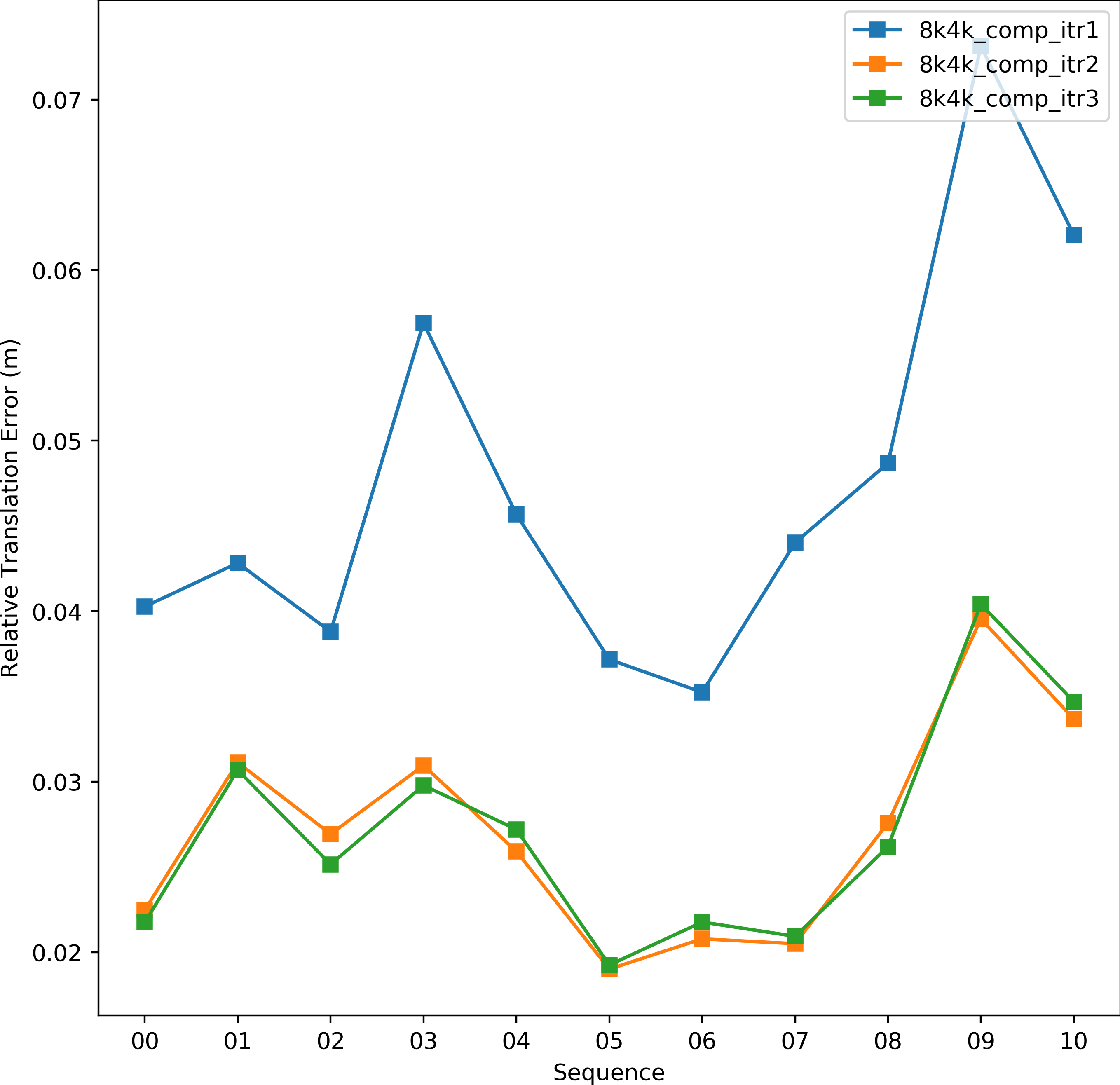}
	\end{center}
	\caption{Comparison of various iterations of the proposed model.}
	\label{fig:itrs}
\end{figure}

In this section we evaluate the performance of the hierarchical model acronymed as \textit{8k4k\_comp}. Figure \ref{fig:trace} shows the trajectories of various training iterations and Figure \ref{fig:itrs} presents their error metrics. As expected, estimates of the second iteration are far superior to the first iteration. The third iteration reduces the error further but the reduction rate is smaller than the previous iteration. This is due to the fact that after certain point, the learning process will plateau as the amount of information to learn is much smaller. The trend is clearly visible in RTE and RRE metrics.



\subsection{Temporal Model}
Another way to increase the accuracy of odometry is by employing temporal features. However, training the frame-to-frame estimation model requires a significant amount of system resources. Adding LSTM on top of that model will dramatically increase these requirements. In such cases, it is common to use a pre-trained feature extraction network and provide mid-level features as inputs to an LSTM model. 
Following this idea, we evaluate three features maps taken from various layers of our trained model.

\begin{itemize}
	\item Feature maps immediately before the flow embedding layer (PREF).
	\item Feature maps immediately after the flow embedding layer (POSF).
	\item Feature maps prior to the fully connected layers in the feature extraction network (FCLF).
\end{itemize}

PREF is the earliest level of features among the three. This model requires the largest temporal model as the flow embedding and the feature extraction layers from the original model are retrained in temporal model.

To reduce the complexity further, POSF features are used. This is achieved by employing features after the flow embedding layer. However, a point-cloud feature extractor needs to be present in the LSTM. 

Finally, FCLF is the smallest feature map. Most of the flow matching and feature extraction work is completed and only the fully connected model is left for the LSTM model.

FCLF features are outperforming both PREF and POSF features. This is the objective that is used to train the frame-to-frame registration model. FCLF mimics the registration model rather than learning temporal features as the provided features are much less informative compared to PREF and POSF.

\begin{figure}
	\begin{center}
		\includegraphics[width=0.49\linewidth]{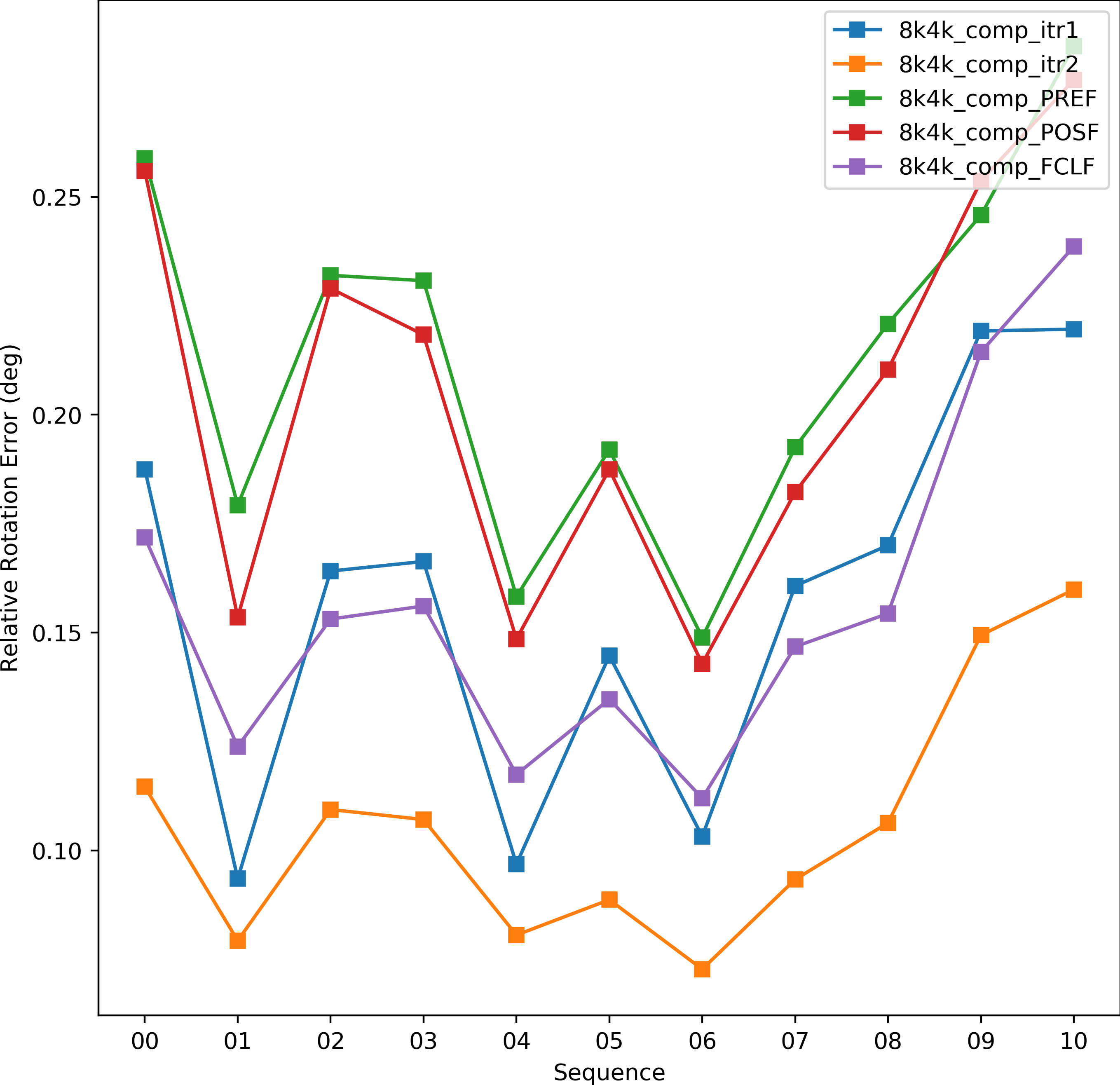}
		\includegraphics[width=0.49\linewidth]{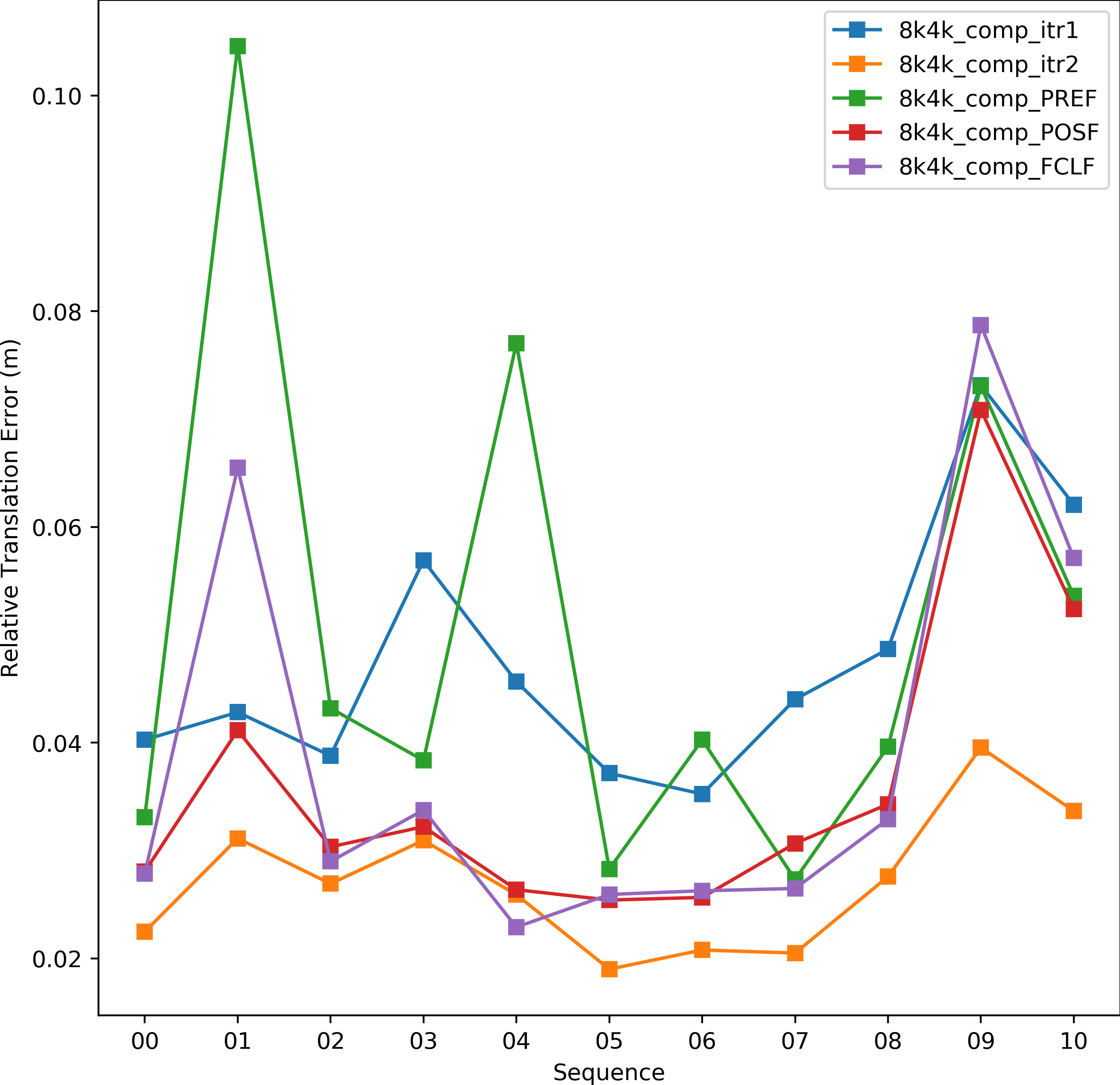}		
	\end{center}
	\caption{Comparison of the performance of the temporal model.}
	\label{fig:lstm}
\end{figure}

Figure \ref{fig:lstm} shows the results of these comparisons. It is seen that the usage of mid-level features is not as fruitful as using a second iteration. In some cases the results are worse than the original feature extraction model that is all attributed to the dividing the training process into two parts.

\subsection{Comparison to the State-of-the-Art}

\begin{table*}[h]
	\centering
	\resizebox{0.98\textwidth}{!}{%
		\begin{tabular}{|c|cc|cc|cc|cc|cc|cc|cc|}
			\hline
			& \multicolumn{2}{c|}{DeepLO} & \multicolumn{2}{c|}{LO-Net} & \multicolumn{2}{c|}{LOAM} & \multicolumn{2}{c|}{$\mathrm{ICP}_{\mathrm{reported}}$} & \multicolumn{2}{c|}{} & \multicolumn{2}{c|}{} & \multicolumn{2}{c|}{} \\
			& \multicolumn{2}{c|}{\cite{cho2019deeplo}} & \multicolumn{2}{c|}{\cite{li2019net}} & \multicolumn{2}{c|}{\cite{zhang2014loam}} & \multicolumn{2}{c|}{\cite{li2019net}} & \multicolumn{2}{c|}{ICP} & \multicolumn{2}{c|}{Ours} & \multicolumn{2}{c|}{Ours+ICP} \\\hline
			Sequence& $t_{rel}$ & $r_{rel}$ & $t_{rel}$ & $r_{rel}$ & $t_{rel}$ & $r_{rel}$ & $t_{rel}$ & $r_{rel}$ & $t_{rel}$ & $r_{rel}$ & $t_{rel}$ & $r_{rel}$ & $t_{rel}$ & $r_{rel}$\\\hline
00       & 0.32 & 0.12   & 1.47 & 0.72   & 1.10  & 0.53   & 6.88  & 2.99   & 32.14 & 13.4    & 3.41 & 1.48   & 2.38 & 1.08 \\\hline
01       & 0.16 & 0.05   & 1.36 & 0.47   & 2.79  & 0.55   & 11.21 & 2.58   &  3.52 & 1.14    & 1.55 & 0.68   & 2.01 & 0.60 \\\hline
02       & 0.15 & 0.05   & 1.52 & 0.71   & 1.54  & 0.55   & 8.21  & 3.39   & 22.64 & 7.02    & 3.35 & 1.40   & 2.81 & 1.27 \\\hline
03       & 0.04 & 0.01   & 1.03 & 0.66   & 1.13  & 0.65   & 11.07 & 5.05   & 37.30 & 4.76    & 6.13 & 2.07   & 5.68 & 1.65 \\\hline
04       & 0.01 & 0.01   & 0.51 & 0.65   & 1.45  & 0.50   & 6.64  & 4.02   &  2.91 & 1.34    & 2.04 & 1.51   & 1.65 & 1.21 \\\hline
05       & 0.11 & 0.07   & 1.04 & 0.69   & 0.75  & 0.38   & 3.97  & 1.93   & 56.91 & 19.93   & 2.17 & 1.09   & 1.74 & 0.99 \\\hline
06       & 0.03 & 0.07   & 0.71 & 0.50   & 0.72  & 0.39   & 1.95  & 1.59   & 29.30 &  6.93   & 2.47 & 1.10   & 1.66 & 0.89 \\\hline
07       & 0.08 & 0.05   & 1.70 & 0.89   & 0.69  & 0.50   & 5.17  & 3.35   & 42.01 & 28.80   & 3.62 & 1.91   & 1.23 & 0.92 \\\hline
08       & 0.09 & 0.04   & 2.12 & 0.77   & 1.18  & 0.44   & 10.04 & 4.93   & 36.58 & 12.36   & 3.61 & 1.61   & 2.74 & 1.29 \\\hline
			$09^{*}$   & 13.35& 4.45   & 1.37 & 0.58   & 1.20  & 0.48   & 6.93  & 2.89   & 36.54 & 12.82   & 8.26 & 3.11   & 2.69 & 1.57 \\\hline
			$10^{*}$   & 5.83 & 3.53   & 1.80 & 0.93   & 1.51  & 0.57   & 8.91  & 4.74   & 28.54 &  6.48   & 11.19& 5.65   & 6.22 & 2.33\\\hline
		\end{tabular}
	}
	\caption{Sequence translation drift percentage and mean sequence rotation error for the lengths of $[100,800]\mathrm{m}$.}
\label{tab:sota}
\end{table*}

In this group of experiments, we compare the proposed model to the state-of-the-art. Table~\ref{tab:sota} compares various models in terms of sequence translation drift percentage and mean sequence rotation error for lengths of $[100,800]\mathrm{m}$. Please note that the results for LOAM~\cite{zhang2014loam}, $\mathrm{ICP}_{\mathrm{reported}}$ are taken from \cite{li2019net}.

LOAM~\cite{zhang2014loam} is one of the benchmarks in this field. It clearly outperforms all of the deep methods in the comparison.
One main reason for that is the feature extraction back-bone network. In our work, we relied on PointNet++~\cite{pointnet++} features. Both LO-Net~\cite{li2019net} and DeepLO~\cite{cho2019deeplo} use 2D depth and surface models such as vertex and normal representations. This way, they completely avoid the usage of 3D data for feature extraction. This trend is also visible across the field as 2D feature extraction models are more advanced than their 3D counterparts. However, there is a growing interest regarding the 3D feature extraction methods \cite{wang2019dynamic}\cite{wu2019pointconv} that can enhance the performance of deep odometry estimation with 3D point clouds.

It is clearly seen that DeepLO~\cite{cho2019deeplo} is over-fitting to the training set. Our model is also suffering from such a phenomenon, but the scale of over-fitting is much lower. One reason for this is the size of our model compared to the size of the input dataset. Using PointNet++ layers results in large networks that require a large amount of data. LO-Net~\cite{li2019net} addresses this problem by utilizing a 2D-based feature extraction network \cite{zhou2017unsupervised}.
We use an implementation of the ICP algorithm from the Open3D library\footnote{\href{www.open3d.org}{http://www.open3d.org/}}. In our implementation we use the sub-sampled point-clouds. This results in a significant drop in performance in comparison to the results of $\mathrm{ICP}_{\mathrm{reported}}$~\cite{li2019net} that use the full point cloud. However, using the full point cloud data entails a large computational complexity burden. The sub-sampling stage that is used to reduce the computational complexity is another aspect that affects the estimation performance of our model. The same points are not always chosen to represent the same static objects. This inherently adds noise to our dataset. We further explore using ICP as a final step on our estimates that significantly improves the performance. This is an expected outcome, as the complexity of the residual problem to solve for ICP at this stage is less than the original one. Hence, it can easily find the corresponding points and estimate better registration parameters.

\subsection{Input Dimensionality}
\begin{figure}[h]
	\begin{center}
		\includegraphics[width=0.49\linewidth]{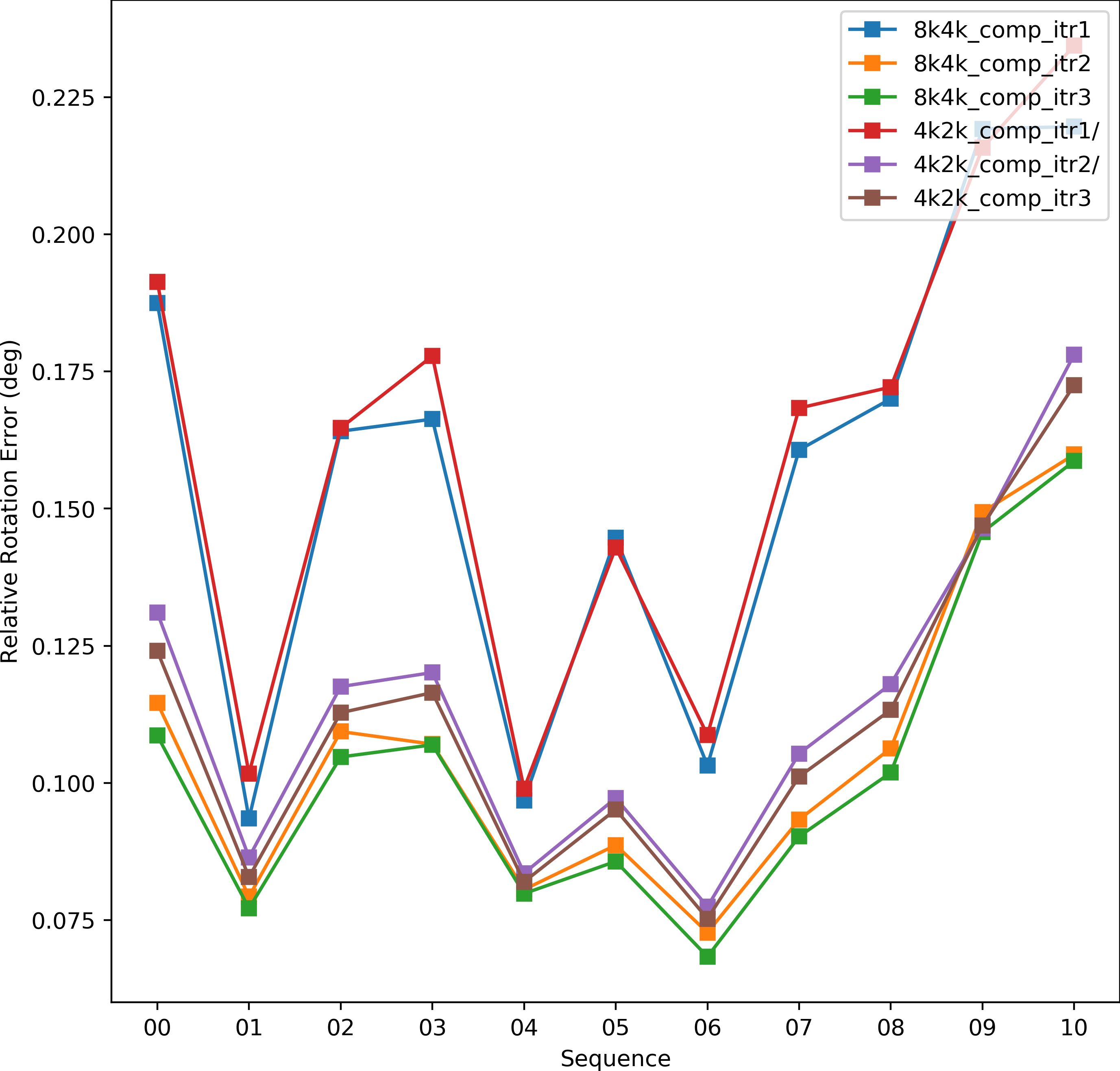}
		\includegraphics[width=0.49\linewidth]{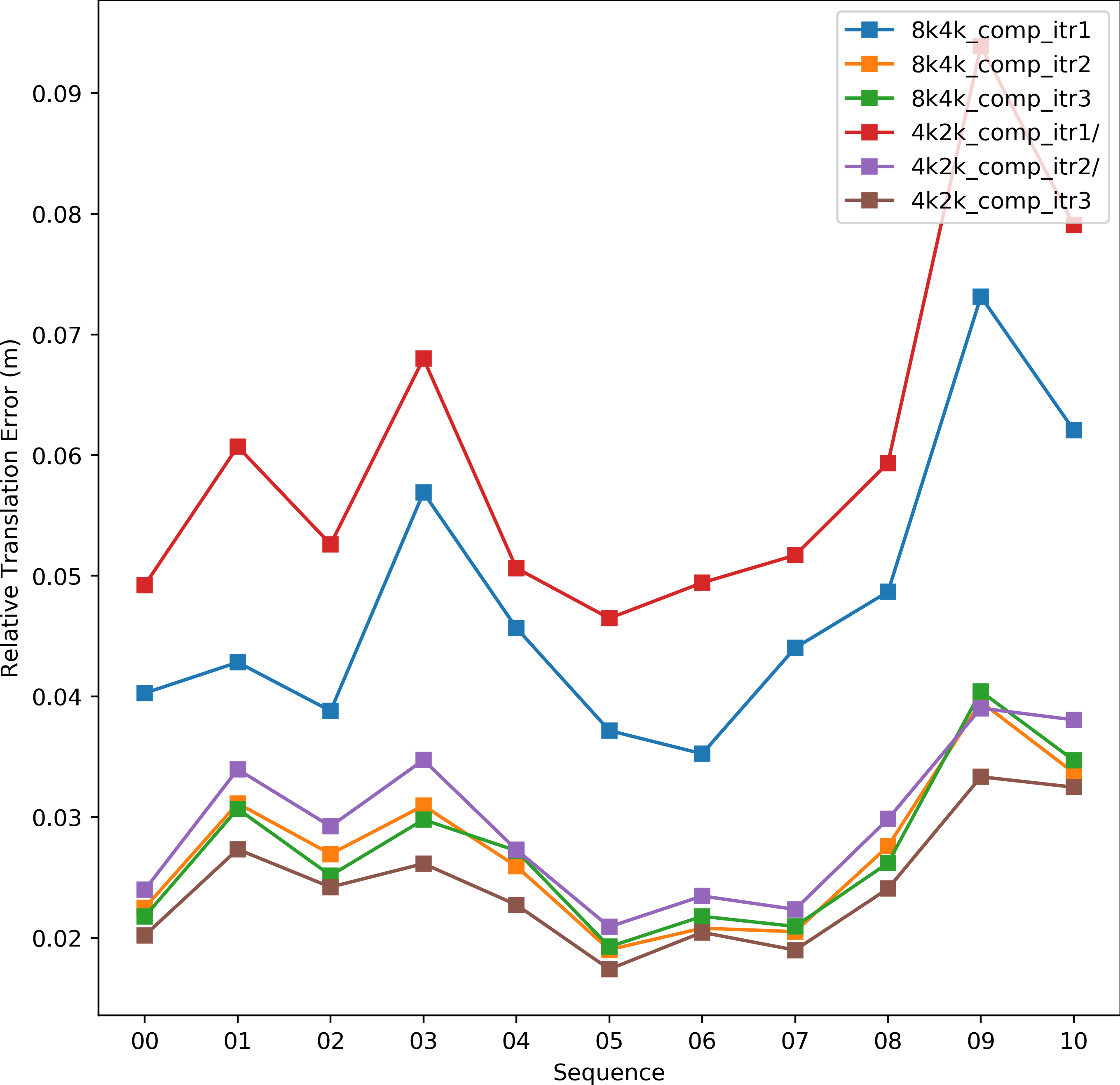}
	\end{center}
	\caption{Input point cloud dimensionality analysis.
		\textbf{\textit{8k4k}} indicates 8k non-ground and 4k ground points in the input cloud.
		\textbf{\textit{4k2k}} uses 4k non-ground and 2k ground points in the input cloud. 
	}
	\label{fig:ablationdims}
\end{figure}

We evaluate the performance of the models with $12k$ and $6k$ input points. \textit{8k4k} represents the 8k and 4k division between non-ground and ground points. Similarly, \textit{4k2k} corresponds to 4k ground and 2k non-ground point sampling. Extra points only help in providing better descriptors at the first layer where the first sampling function in the network is called. This results in better performance of the model, especially in the first iteration where the disparity between matching points in two frames is much larger. However, the difference diminishes in the second and third iterations. This entails that by employing a hierarchical model we could reduce the complexity of the input point cloud by using coarser 3D point clouds. Results of this comparison are shown in Figure~\ref{fig:ablationdims}. 

\subsection{Sampling Comparison}
\begin{figure}[h]
	\begin{center}
		\includegraphics[width=0.49\linewidth]{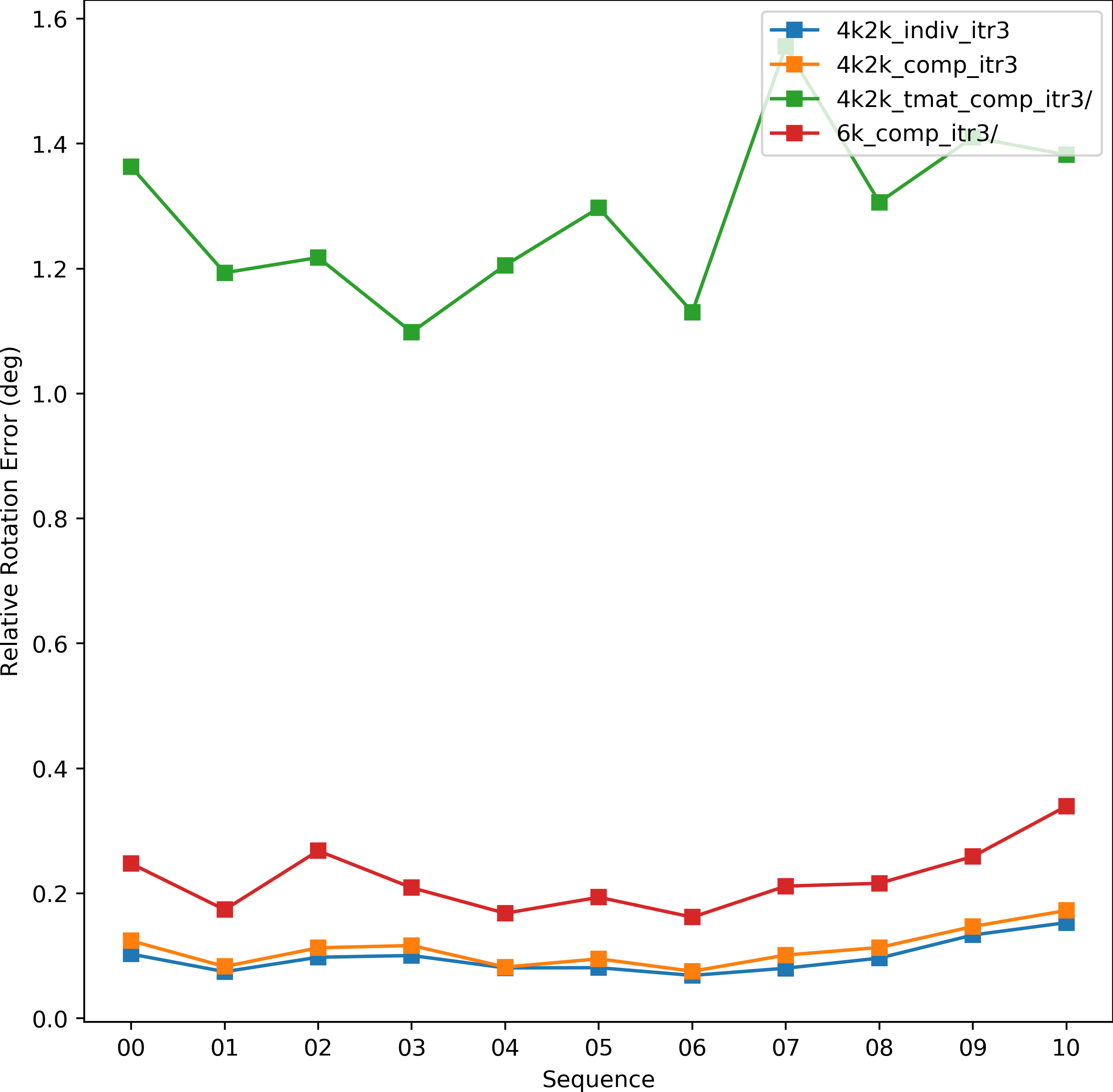}
		\includegraphics[width=0.49\linewidth]{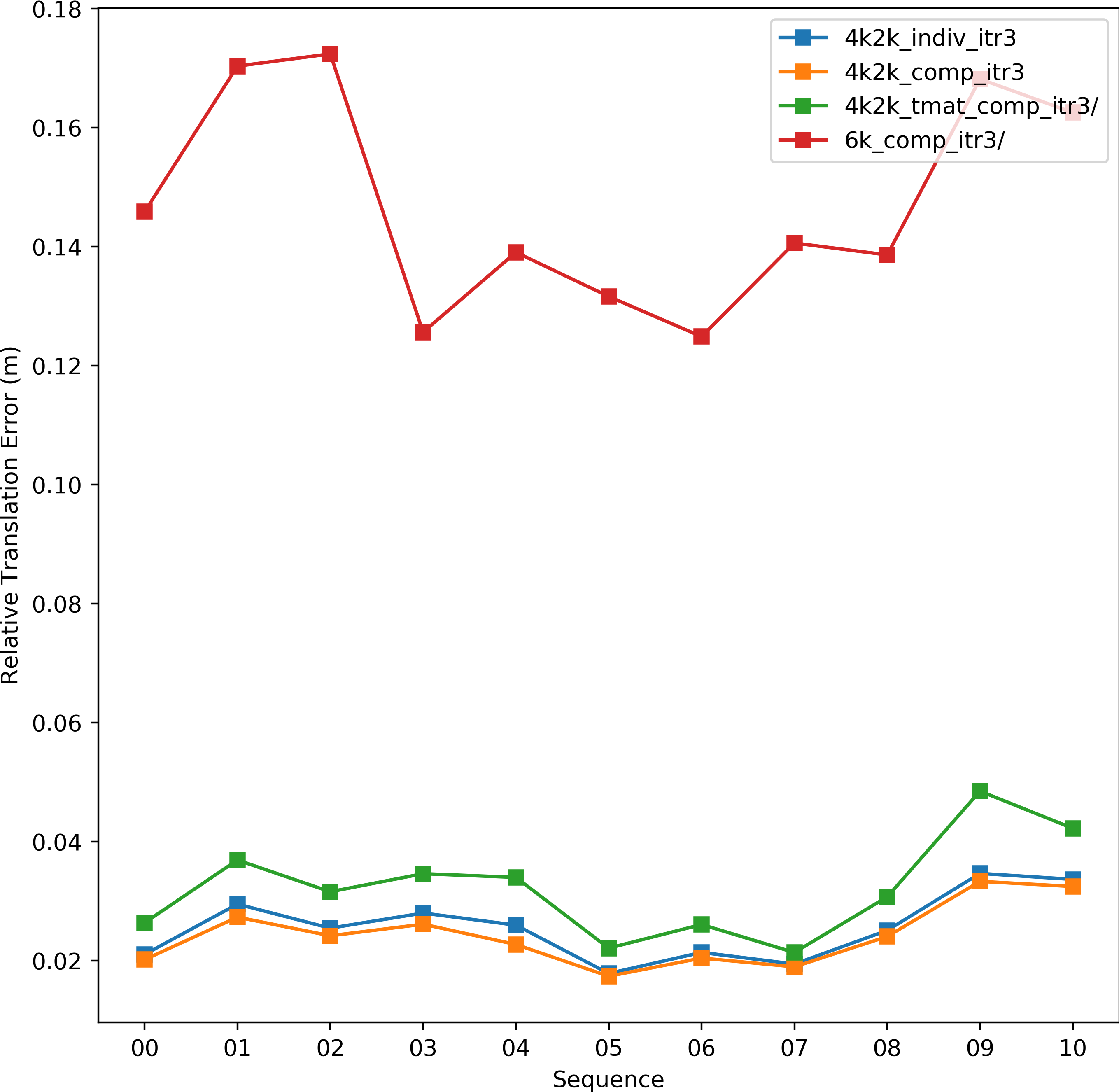}
	\end{center}
	\caption{Comparative results for various ablation studies.
		\textbf{\textit{4k2k}} uses 4k non-ground and 2k ground points as its input. 
		\textbf{\textit{6k}} indicates that 6k points are globally sampled without any distinction between ground and non-ground points.
		\textbf{\textit{comp}} uses one weight for the rotation component and one weight for the translation component in the loss function.
		\textbf{\textit{indiv}} uses individual weights for each parameter in the pose.
		\textbf{\textit{tmat}} indicates that the rotation matrix is used instead of normalized Euler angles to calculate the loss.
	}
	\label{fig:ablation}
\end{figure}

To better understand the importance of separately sampling ground and non-ground points, we train the same model (\textit{6k\_comp}) with $6k$ globally sampled points from the point cloud. We train this model in hierarchical manner and compare it to the \textit{4k2k\_comp} model that employs 4k non-ground and 2k ground input points. As it is shown in Figure~\ref{fig:ablation}, sampling without distinction between ground and non-ground points results in far worse performance. This is due the large number of ground points in the point cloud that provide much less information regarding translation and orientation of the sensor in comparison to the non-ground points.

\subsection{Loss comparison}
The proposed model employs weighted \textit{l2\_norm} loss on Euler angles and translation parameters. The loss function utilizing 2 weights for each rotation and translation components is indicated with \textit{comp}, while the \textit{indiv} represents the usage of individual weights for each transformation parameter. 

\begin{figure}[h]
	\begin{center}
		\includegraphics[width=0.49\linewidth]{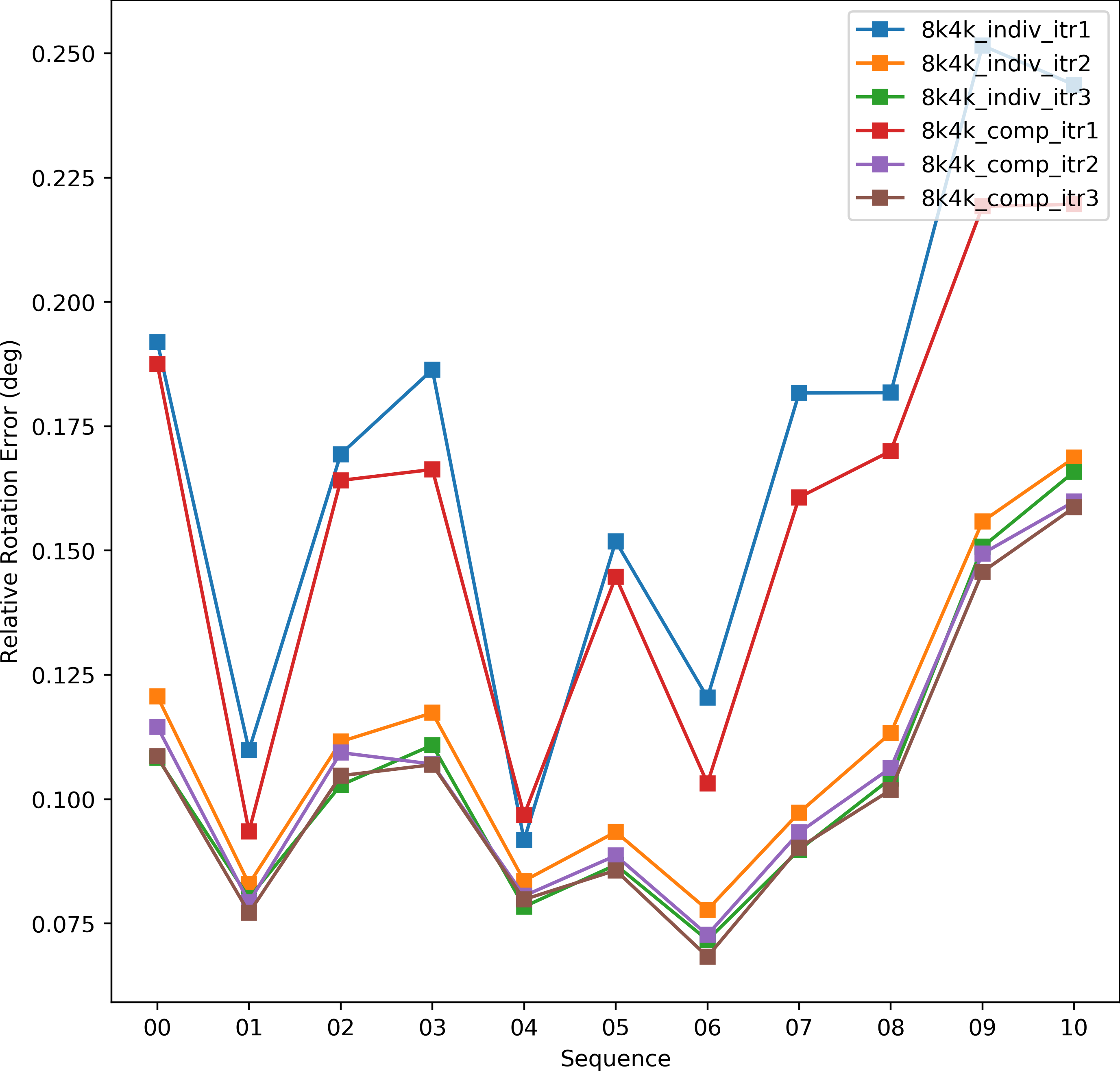}
		\includegraphics[width=0.49\linewidth]{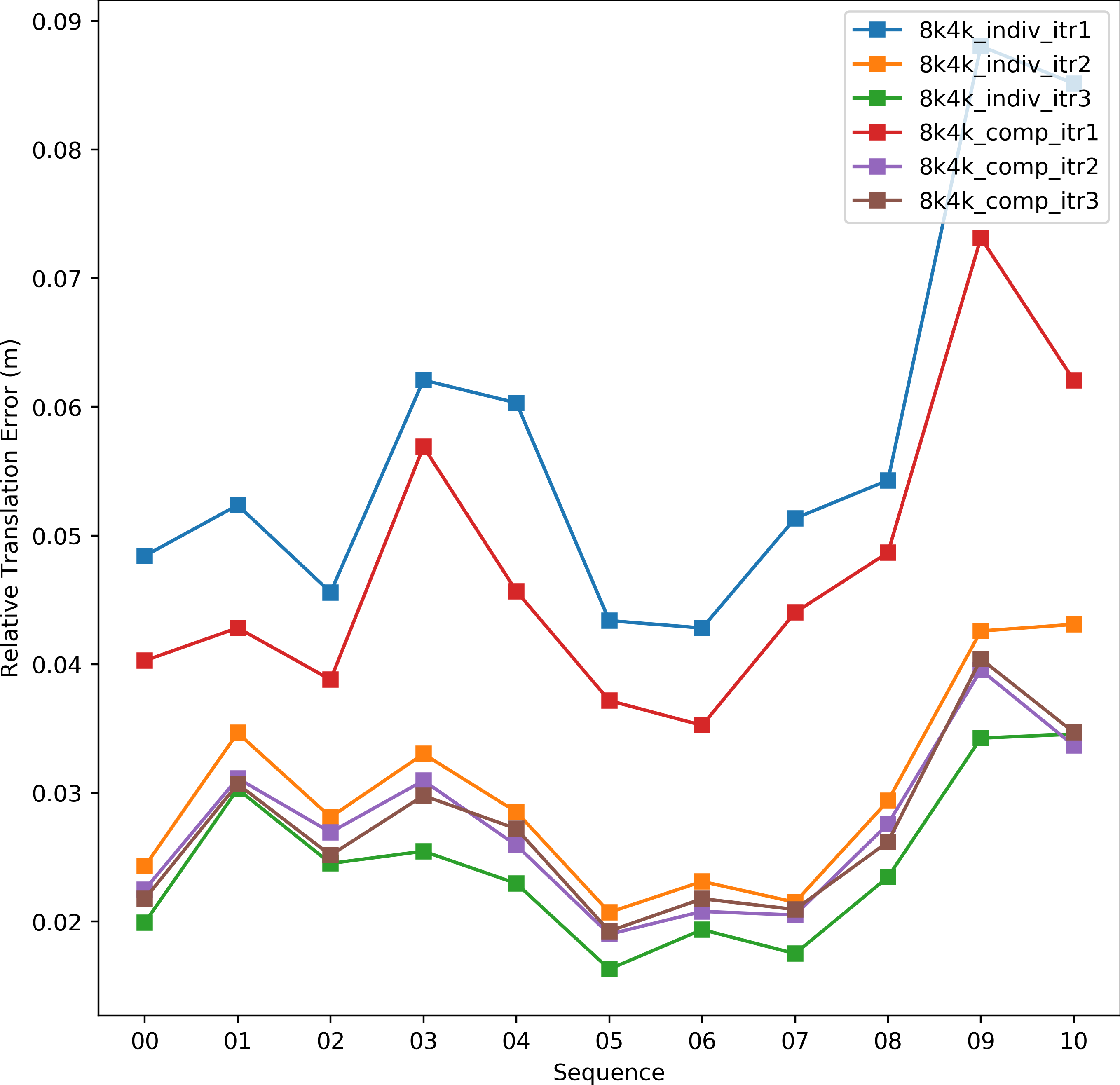}
	\end{center}
	\caption{Comparative results over various iterations for \textit{comp} vs \textit{indiv} weighting scheme in loss function.
		\textbf{\textit{comp}} uses one weight for rotation component and one weight for translation component in the loss function.
		\textbf{\textit{indiv}} uses individual weights for each parameter in the pose.
	}
	\label{fig:ablation_compindiv}
\end{figure}

Figure~\ref{fig:ablation_compindiv} shows comparative results of this experiment. We observe that in the first two iterations, component based weighting provides better results. However, in the third iteration, individual weighting achieves comparable results to the component based function. It is worth mentioning that the scale of reduction in error between iteration 2 and 3 is small, and the majority of the error reduction is achieved in the first 2 iterations.

\subsection{Label Representation}
Normalized Euler angles are used as the primary labels along with normalized translation parameters in our experiments. Normalization is utilized in order to remove the scaling effects for various parameters in calculation of the \textit{L2-norm}. To validate our decision, we compare our choice of label representation to the transformation matrix representation with 12 parameters ($3\times3$ rotation and $3$ translation). We employ the component-based weighting on rotation and translation components of this representation. The trained model is shown as \textit{4k2k\_tmat\_comp} in Figure~\ref{fig:ablation}.
Results show that normalized Euler angles are a much better representation than the $3\times3$ rotation matrix, which is an over-parameterized representation of the rotation.

\section{\uppercase{Conclusion}}
\label{conclusion}
In this paper, we have proposed a methodology to use deep neural networks to estimate odometry based on 3D point clouds. We have proposed a data augmentation mechanism along with measures incorporated in the loss function to estimate the frame-to-frame transformation parameters. The proposed model successfully reduces the error in consecutive iterations. Furthermore, we have evaluated the usage of pre-trained feature maps for training temporal models. Our results are comparable to the state-of-the-art. We argue that the extracted features from the 3D point clouds are not descriptive enough for this task. 3D point-cloud-based deep learning is still a new field and 3D deep feature extraction techniques have not matured as much as their 2D image-based counterparts. 


\bibliographystyle{apalike}
{\small
\bibliography{example}}

\end{document}